\documentclass[journal]{IEEEtran}

\ifCLASSINFOpdf
\else
\fi
\usepackage{amssymb}
\usepackage{graphicx}
\usepackage{amsmath}

\usepackage{url}
\urldef{\mailsa}\path|{alfred.hofmann, ursula.barth, ingrid.haas, frank.holzwarth,|
	\urldef{\mailsb}\path|anna.kramer, leonie.kunz, christine.reiss, nicole.sator,|
	\urldef{\mailsc}\path|erika.siebert-cole, peter.strasser, lncs}@springer.com|    

\usepackage{color}

\usepackage{array}
\newcolumntype{x}[1]{>{\centering\arraybackslash\hspace{0pt}}p{#1}}
\usepackage{multirow}
\usepackage{bbding}
\usepackage{fixltx2e}
\usepackage[hidelinks]{hyperref}% http://ctan.org/pkg/hyperref

\newcolumntype{L}[1]{>{\raggedright\let\newline\\\arraybackslash\hspace{0pt}}m{#1}}
\newcolumntype{C}[1]{>{\centering\let\newline\\\arraybackslash\hspace{0pt}}m{#1}}
\newcolumntype{R}[1]{>{\raggedleft\let\newline\\\arraybackslash\hspace{0pt}}m{#1}}

\usepackage[extendedchars]{grffile}

\usepackage{comment}
\usepackage{xspace}
\def\eg{\textit{e.g.}~}

\def\etal{\textit{et al.}\xspace}
\def\ie{\textit{i.e.}~}
\def\cf{\textit{cf.}~}

\def\wrt{w.r.t. }

\begin{document}
	
\title{Weakly-Supervised Localization and Classification of Proximal Femur Fractures}
	%Insights on Understanding Deep Learning Models
	
	
	\author{Amelia Jim\'{e}nez-S\'{a}nchez$^{*}$\href{https://orcid.org/0000-0001-7870-0603
}{\includegraphics[scale=0.5]{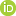}}, 
	Anees Kazi$^{*}$, Shadi Albarqouni\href{https://orcid.org/0000-0003-2157-2211}{\includegraphics[scale=0.5]{images/orcid.png}}, Sonja Kirchhoff, Alexandra Str{\"a}ter, Peter Biberthaler, Diana Mateus$^{**}$, and~Nassir Navab$^{**}$	
	
	%\author{Amelia Jim\'{e}nez-S\'{a}nchez$^{*}$, 
	%Anees Kazi$^{*}$, \IEEEmembership{Student Member, IEEE}, Shadi Albarqouni, \IEEEmembership{Member, IEEE}, Sonja Kirchhoff, \IEEEmembership{MD}, Alexandra Str{\"a}ter, \IEEEmembership{MD}, Peter Biberthaler, \IEEEmembership{MD}, Diana Mateus$^{**}$, 
		%and~Nassir Navab$^{**}$, \IEEEmembership{Member, IEEE}% <-this % stops a space
		\thanks{* A. Jim\'{e}nez-S\'{a}nchez and A. Kazi have equally contributed to the work. Sequence of names are based on alphabetic order.}% <-this % stops a space
		\thanks{** D. Mateus and N. Navab are joint senior authors.}
		\thanks{A. Kazi and S. Albarqouni are with Computer Aided Medical Procedures (CAMP), TU Munich, Garching, Germany.
		}% <-this % stops a space
		\
		\thanks{A. Jim\'{e}nez-S\'{a}nchez was with CAMP, TU Munich, Germany during the work. She is currently with Department of Communication and Information Technologies (DTIC), Universitat Pompeu Fabra, Barcelona, Spain.}
		\thanks{S. Kirchhoff is with the Dept of Trauma Surgery, Klinikum rechts der Isar, TU Munich, Munich, Germany.}
		\thanks{A. Str{\"a}ter is with Dept of Diagnostic and Interventional Radiology, Klinikum rechts der Isar, TU Munich, Munich, Germany.}
		\thanks{P. Biberthaler is with the Institute of Clinical Radiology, Ludwig Maximilian University, Munich, Germany, and Head Dept. of Trauma Surgery, Klinikum rechts der Isar, TU Munich, Munich, Germany.}
		\thanks{D. Mateus is with LS2N, CNRS UMR 6004, Ecole Centrale Nantes, France.}
	    \thanks{N. Navab is with CAMP, TU Munich and Johns Hopkins University, Baltimore MD, USA.}}


	\markboth{}
	{Jim\`{e}nez-S\`{a}nchez \MakeLowercase{\textit{et al.}}}



	\maketitle
	
	%\include{abstract}
	%pixel-level
	\begin{abstract}
	
In this paper, we target the problem of fracture classification from clinical X-Ray images towards an automated Computer Aided  Diagnosis (CAD) system. Although primarily dealing with an image classification problem, we argue that localizing the fracture in the image is crucial to make good class predictions. Therefore, we propose and thoroughly analyze several schemes for simultaneous fracture localization and classification. We show that using an auxiliary localization task, in general, improves the classification performance. Moreover, it is possible to avoid the need for additional localization annotations thanks to recent advancements in weakly-supervised deep learning approaches. Among such approaches, we investigate and adapt Spatial Transformers (ST), Self-Transfer Learning (STL), and localization from global pooling layers. We provide a detailed quantitative and qualitative validation on a dataset of 1347 femur fractures images and report high accuracy with regard to inter-expert correlation values reported in the literature. Our investigations show that i) lesion localization improves the classification outcome, ii) weakly-supervised methods improve baseline classification without any additional cost, iii) STL guides feature activations and boost performance. We plan to make both the dataset and code available.
	\end{abstract}


	% \fi
	\IEEEpeerreviewmaketitle
	% !TEX root = ../AutoClassPFF_JBHI.tex

	\section{Introduction}
	%\IEEEPARstart{F}{ractures} 
	% MEDICAL MOTIVATION
	\IEEEPARstart{F}{ractures} of the human skeleton, especially of the long bones, are among the most common reasons why patients visit the emergency room. Initial diagnostics include X-rays of the affected bone, mostly from two orthogonal projections. X-rays are evaluated by either expert radiologists and/or trauma surgeons for the classification of the fracture, resulting in different treatment options. Most commonly, the fracture is classified according to the guidelines of the Arbeitsgemeinschaft Osteosynthese (AO). However, for certain bones, several classification guidelines exist with no agreement in the traumatology community \cite{garnavos2012new,newton1985etiology}.
    %swiontkowski1997interobserver,muller2012comprehensive}.
	
Accurate fracture classification requires years of experience, as reflected by the reported inter-observer variability, which is as low as 66$\%$ for residents and 71$\%$ for experts~\cite{vanEmbden:injury2010}. 
The aim of our study is, therefore, to predict the fracture type on the basis of digital radiographs to assist physicians especially young residents and medical students.  We primarily focus on predicting fracture type of the proximal femur according to the AO classification, for which a good reproducibility was published~\cite{Jin}.

\begin{figure}
		\centering
		\includegraphics[width=0.9\linewidth]{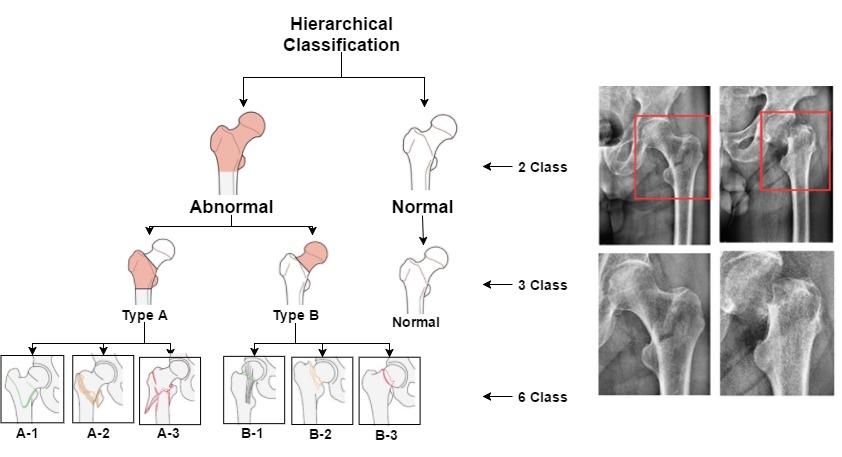}
		\caption{(left) Hierarchical AO fracture classification into 2, 3 and 6 classes. (right) Example digital X-ray showing the fracture localized within a bounding-box (top) and the zoomed-in regions of interest (bottom).}
		\vspace{-0.5cm}
		\label{fig:aotree}
	\end{figure}

For most anatomical locations,  the AO classification is hierarchical and depends upon localization and configuration of the fracture lines. In the top level of the hierarchy as shown in Fig.~\ref{fig:aotree}, one searches whether a fracture is present in the imaged anatomy. Then, the fracture is classified as type A or B according to localized information. 
Finally, finer grained classes are considered (A1, A2, ..., B3)  according to the number of fragments and their respective displacement. 

	\begin{comment}
	% CHALLENGES, PRIOR WORK our APPROACH
	\begin{figure}
		\centering
		\includegraphics[width=0.9\linewidth]{images/Challanges}
		\caption{\textbf{top} Example images with classification label showing the challenges such as low contrast, clutter.\textbf{bottom} Cropped regions of interest.}
		\vspace{-0.5cm}
		\label{fig:ao}
	\end{figure}
	\end{comment}
	
Automatic classification of fractures from X-ray images is confronted with several challenges. From an image analysis point-of-view, fractures appear with low SNR and poor contrast, they have unpredictable shapes, and may resemble and superpose to other structures on the image (see Fig.~\ref{fig:aotree}). Approaches for automatic fracture classification tackling these challenges have appeared only recently relying mostly on intermediate feature extraction steps~\cite{burns:radiology2016:spineCT,wu:jbioim2012:fractureCT,ebsim:clip2016:miccaiwshp,bayram:2016diffract}. 

From a broader perspective, determining the type of fracture on the basis of X-rays is an instance of a medical image classification problem, for which many of the state-of-the-art solutions are based on Deep Neural Networks (DNN)~\cite{roth2016deep}. 
Very recently, the classification of full X-ray images with well-known DNN architectures has delivered promising results~\cite{olczak2017artificial}. In this paper, we argue that the localization of the region of interest (ROI), containing the fracture lines and its immediate surrounding, is crucial and better suited than the full image for AO fracture classification. Our motivation comes from two observations: first, the fracture represents only a small portion of the image, and second,  the fine-grained classification depends upon information only visible at high resolutions. 

In order to solve the localization task, one can resort to supervised localization methods, which perform well but increase the need for time-consuming expert annotations. We favor therefore weakly-supervised schemes where the classification task is supervised but the localization task is trained only from the class annotations at the image level. 
Our initial approach in this direction was presented in~\cite{kazi2017automatic}, was based on localization of the ROI via Spatial Transformers (ST) networks~\cite{jaderberg2015spatial} with promising results.
In this study our  contributions are: 
	\begin{enumerate}
		\item Formally studying the hierarchical classification of proximal femur fractures, using four types of deep learning schemes: with and without localization, and with supervised / weakly-supervised localization. 
		\item Investigating various state-of-the-art \textbf{\textit{weakly-supervised}} methods, aiming to improve the classification performance with the help of localization but without additional localization annotations. \\
$\bullet$ We model the problem in the framework of deep attention models~\cite{jaderberg2015spatial,girshick:cvpr2014:rcnn}, capable of finding a ROI from the image content. \\
$\bullet$ We take advantage of  Self-Transfer Learning (STL) \cite{hwang2016self} methods, which jointly train localization and classification as one task. \\
$\bullet$ We provide a comparative study of the influence of different pooling schemes (average, max and Log-Sum-Exponential (LSE) pooling \cite{wang2017chestx}) on the classification and localization tasks. 
\item Presenting classification and localization results for 2, 3 and 6 classes on a dataset of 672 patients. We further discuss techniques to handle class imbalance: modified weighted scheme for cross-entropy loss.
	\end{enumerate}
	
\subsection{Related Work}

The problems of automatic bone fracture detection and classification have  first been addressed in the literature with conventional \textit{machine learning}  pipelines consisting of preprocessing, feature extraction and classification steps. Preprocessing steps typically involve noise removal and edge detection~\cite{al2013detecting} or thresholding. Feature extraction methods go from the general texture features like wavelets \cite{al2013detecting} or Gabor \cite{7163993} to application-specific ones like  the bone completeness indicator and the fractured region mapping \cite{bayram:2016diffract}. Finally, the leveraged classification methods  include Random Forests (RF)~\cite{7163993}, BayesNet~\cite{al2013detecting} and Support Vector Machines (SVM)~\cite{lim2004detection}.

Within the past two years and with the advent of \textit{deep learning models} several approaches toward computer-aided decisions related to fractures have been proposed. Particular interest has been given to the detection of spine fractures in CT images. For instance, Roth~\etal~\cite{roth2016deep} propose the use of a Convolutional Neural Network (ConvNet) trained to predict a probability map of fracture incidence. Also, Bar~\etal~\cite{bar2017compression} describe a method relying on a ConvNet followed by a Recurrent Neural Network (RNN) for compression fracture detection. 
Interestingly, the two methods above use patch-wise training of the networks. This is an indicator of the need to preserve the resolution to a level of detail where the fracture details are visible. However, the lesion localization is done either by analyzing every patch independently~\cite{roth2016deep}, or with a pseudo-automated aproach~\cite{bar2017compression} based on first detecting and scrolling over the spine. A patch-wise approach implies an additional level of decision to determine the class of the fracture. Instead, our efforts focus on strategies that automatize the localization of the ROI. By doing so, we also reduce the need for pixel-wise detection or bounding-box annotations used in~\cite{roth2016deep,bar2017compression}. While the two methods above require pre-processing steps for either edge extraction and alignment or spine segmentation, we opt for end-to-end solutions.

There are only a few methods that target the fracture classification task from X-ray acquisitions, where edges of fractures may be superposed to other structures.  Based on bone segmentation, high-level features and RF, Bayram~\etal~\cite{bayram:2016diffract} proposed a method to classify femur shaft fractures. Although the application is similar to ours, classification of the shaft is simpler as less clutter is present. 
A contemporary work to ours is that of Olczak~\etal~\cite{olczak2017artificial} who aims at investigating the behavior of different deep models for various tasks such as classification of body parts, X-ray views, fractures/normal, dexter/sinister. While the method uses the full-image as input to the different deep models, we demonstrate here the benefits of first localizing the ROI. Finally, both X-ray classification methods~\cite{bayram:2016diffract, olczak2017artificial}, targets the 2 class (normal/ abnormal) classification problem, whereas we target the more challenging fine-grained classifications of the AO standard.

A popular method for ROI localization is Regions with CNN (R-CNN) \cite{girshick:cvpr2014:rcnn}, which however depends on an external region proposal system.
In order to localize the ROI without the need of additional expert annotations,  we model the problem in the framework of deep attention models \cite{jaderberg2015spatial,girshick:cvpr2014:rcnn} capable of finding a ROI implicitly. In addition, we leverage STL \cite{hwang2016self}, which optimizes simultaneously for the classification and localization tasks.
\vspace{-0.4cm} 
	\section{Methodology}
\label{sec:method}
\def\params{{\bf p}}
\def\im{{\bf I}}
\def\crop{\im'}
\def\lab{y}
\def\pred{\hat{y}}
\def\loc{g}
\def\locw{\omega_g}
\def\class{f}
\def\classw{\omega_f}
\def\warp{\mathcal{W}_{\params}}
\def\real{\mathbb{R}}
\def\grid{\mathcal{G}}

\begin{figure*}[t]
\centering
	\includegraphics[width=0.8\textwidth]{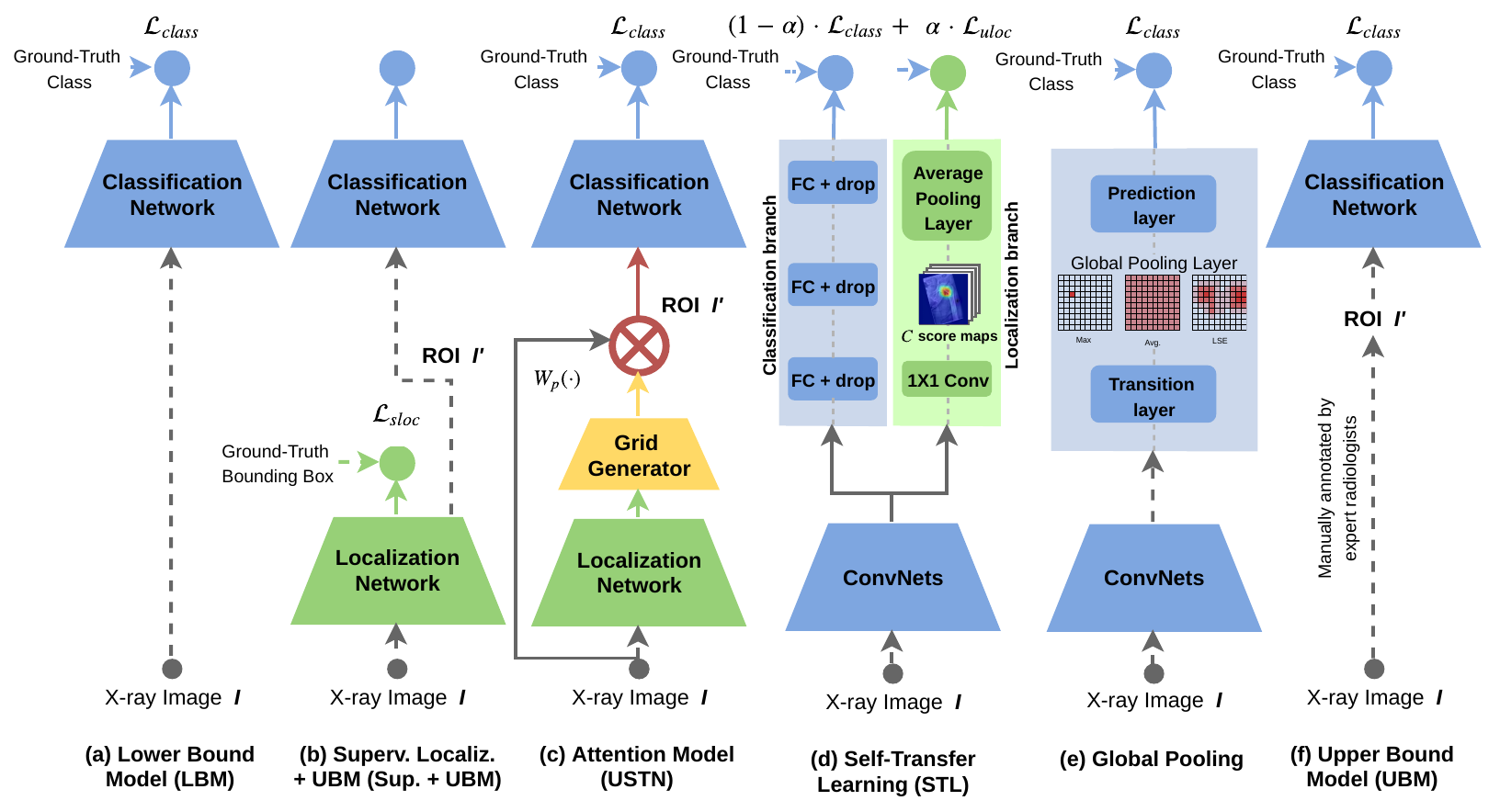}
	%\caption{Schematic representation of all the considered models.}
	\caption{Schematic representation of all the considered models. First (a) and last (f) models are the lower- and upper-bound references, respectively. (b) Combination of supervised localization and classification of the predicted ROI. (c) Model based on STN, learns implicitly the transformation to predict the bounding box that is classified. (d) STL consists of a classification and localization branch, which aims to improve both. With model (e) we investigate how different pooling schemes influence the performance.}
	\label{fig:overview}
	\vspace{-.5cm}
\end{figure*}

Given $N$ X-ray images with each image $\im \in \real^{H\times W}$, our aim is to build a {\it classification model} $\class(\cdot)$ that assigns to each image a class label $\lab \in C$, where $C \subset \rm \{{normal, A1,A2,A3,B1,B2,B3}\}$, \ie  
$\pred = f(\im; \classw)$, where $\hat{}$ denotes a prediction and  $\classw$ are the classification model parameters. In addition, we define a {\it localization task} $g(\cdot)$ that returns the position $\params$ of the ROI, $\crop$, within the X-ray image such that
$\hat{\params} = g(\im; \locw)$,
where $\locw$ are the localization model parameters. $\params = \{t_{r}, t_{c}, s\}$ is a bounding box of scale $s$ centered at $(t_{r}, t_{c})$. The ROI image $\crop\in\real^{H'\times W'}$ is obtained with 
$\crop = \warp(\im)$,
where $\warp(\cdot)$ is a warping operator. In the following, we detail the different approaches to solve the classification and localization tasks.

\subsection{Automatic Classification}
\label{ssec:AC}
In this subsection, we define  the lower- and upper-bound models of the classification task. They will serve as reference to measure the influence of localization of the ROI on the classification accuracy. 
We take as {\it Lower-bound} (LBM) the model making a fracture-type prediction given the whole X-ray image as input, \ie $\hat{y} = f(\im; \classw)$. Conversely, the {\it Upper-bound Model} (UBM) is defined as the classification of a manually annotated ROI, \ie $\hat{y} = f(\crop; \classw)$. The schematic representation of both LBM and UBM are depicted in Fig.~\ref{fig:overview}a, and Fig.~\ref{fig:overview}f, respectively. 

We use independent ConvNets to approximate the LBM and UBM.
To handle the large class imbalance present in the dataset,  
both models are optimized to minimize the weighted cross-entropy (WCE) loss function:
\begin{equation} \label{eq:wce}
\mathcal{L}_{class} = - \sum_{j\in C} w_c \cdot y_j\log(\hat{y_j}),
\end{equation}
where $w_c$ is a class-associated weight that penalizes errors in the under-represented classes and computed using the reciprocal of the class frequency as:
\begin{equation}
w_c = \frac{Z}{{\rm freq}(c)},  
\end{equation}
with $Z$ a normalization coefficient. When $w_c = 1$, the loss function is the standard cross-entropy (CE).
Notice that the localization task was not considered here. 

\subsection{Supervised Localization and Classification (Sup. + UBM):} 
In our third approach, we model the  localization  $g(\cdot)$ and classification $\class(\cdot)$ as independent tasks (see 
 Fig.~\ref{fig:overview}b). The model for classification is equivalent to the UBM from the previous section, while 
$g(\cdot)$  is modeled with a regression ConvNet minimizing the loss:
\begin{equation}
\mathcal{L}_{sloc} = \frac{1}{2}\| \params - \hat{\params}\|^2, 
\label{eq:l2}
\end{equation}
where $\|\cdot\|$ is the $\ell_2$-norm, and $\hat{\params}$ is the predicted bounding box. The output localized ROI image is then fed to $f(\crop; \classw)$, only for evaluation. 

\subsection{Weakly-Supervised Models} 
In this subsection, we address the localization task $g(\cdot)$ when no bounding box annotations are available. We refer to these models as weakly-supervised, as they still require the supervision of the fracture-class. We adapted attention, self-transfer learning and global pooling approaches.

\subsubsection{Attention Model}
As presented in~\cite{kazi2017automatic} and depicted in Fig.~\ref{fig:overview}c, we propose training $g(\cdot)$ and $\class(\cdot)$ tasks jointly. To deal with the lack of supervision in the localization, we follow the principles of STN~\cite{jaderberg2015spatial} to implicitly learn the parameters of the transformation $\warp(\cdot)$.
Based only on class annotations, STN searches for the ROI which improves the classification. 
In practice, STN  regresses the warping parameters $\params$ from the back-propagation of the classification loss $\mathcal{L}_{class}$ through a localization network.
Since locally sharp differences in the class prediction will appear only at the fracture location, STN will tend to localize the femur head. In practice, STN is trained to regress the warping parameters $\params$ from the back-propagation of the classification loss. For further details, readers are referred to~\cite{jaderberg2015spatial}.

The warping $\warp(\cdot)$ is defined by a transformation matrix $\mathcal{T}_{\theta_p}$ and a bilinear interpolation allowing for cropping, translation, and isotropic scaling.  $\warp(\cdot)$ samples the locations of the ROI grid $\{x_i',y_i'\}_{i=1}^{\rm gridSize}\in\grid$ from the original image:
\begin{equation}
\left[ \begin{array}{c} x_i \\ y_i \end{array} \right] = 
\warp
\left( \left[ \begin{array}{c} x'_i \\ y'_i \end{array} \right] \right) = \mathcal{T}_{\theta_p} \cdot \left[ \begin{array}{c} x'_i \\ y'_i \end{array} \right],
\end{equation}
where $x_i$ and $y_i$ are the mapped locations of the grid in the input image $\im$, and $\theta_p$ are the transformation parameters. 
This setting is referred as Unsupervised Spatial Transformer Network (USTN). We study two types of matrix transformation: i) a similarity transform (a-STN) with three parameters and ii) affine transformation (aff-STN) for six parameters.

\subsubsection{Self-Transfer Learning (STL)}
Inspired by Kim \etal~\cite{hwang2016self}, we modeled the localization $g(\cdot)$ as an auxiliary task  trained jointly with $\class(\cdot)$ . The STL framework, as depicted in Fig.~\ref{fig:overview}d, consists of shared convolutional layers, followed by two branches: one for classification and the other for localization. Their role and significance are described below:
\begin{itemize}
	\item \textit{Shared convolutional layers}: The model at first comprises  $n$ shared convolutional layers as shown in Fig.~\ref{fig:overview}d. The input given to this shared block are images of slightly larger dimensions (500x500 px in our case) than expected for a conventional ConvNet, to create larger activation maps. These bigger activation maps give a better insight to the pooling layer, which through back-propagation generates $C$ different activation maps highlighting the location of $C$ class objects. %
	\item \textit{Classification branch:} It is modeled as $f(\im; \classw, \omega_s)$ using fully connected layers and dropout. Both the weights of the fully connected layers $\classw$ and the shared convolutional layers $\omega_s$ minimize the cross-entropy loss $\mathcal{L}_{class}$. 
	\item \textit{Localization branch}: It is modeled as $g(\im; \locw, \omega_s)$ using convolutional layers to bring the output activation maps of the shared convolutional layers to $C$ activation (score) maps followed by an average global pooling layer. Both the weights of the convolutional layers $\classw$ and the shared convolutional layers $\omega_s$ are learned to minimize the cross-entropy loss $\mathcal{L}_{uloc}$ as 
	\begin{equation}
	\mathcal{L}_{uloc} = - \sum_{j\in C} y_j\log(g(\im; \locw, \omega_s)).
	\label{eq:uloc-loss}
	\end{equation}   
	This branch help us find the localization of the ROI without the need for localized information.
	\item \textit{Joint loss function and hyper-parameter $\alpha$}: Both classification and localization branches are trained together in an end-to-end way to minimize the following loss function:
	\begin{equation}
	\mathcal{L}_{total} = (1-\alpha) \cdot \mathcal{L}_{class} + \alpha \cdot \mathcal{L}_{uloc},
	\label{eq:stl-loss}
	\end{equation}
	where $\alpha$ is the hyper-parameter controlling the contribution of each of the branches during training. The value of $\alpha$ is initially kept small allowing training the filters of shared layers for better classification accuracy. When reaching convergence, the value of $\alpha$ is complemented, i.e. set to $1 - \alpha$, to focus on improving the localizer. 
\end{itemize}
Finally, with such alternate training the model learns to target and extract relevant features from the ROI. This boosts the performance of our primary classification task. 

\subsubsection{Global Pooling}
Inspired by Wang~\etal~\cite{wang2017chestx}, the {\it classification task} $f(\im; \classw)$ is modeled using pre-trained ConvNets models followed by a Transition layer, a Global pooling layer, and a Prediction layer without the need of an auxiliary {\it localization task} (see Fig.~\ref{fig:overview}e). The ROI localization is then obtained using a simple convolution operation between the output of the transition layer and the learned weights of the prediction layer. We describe the role of the layers below:
\begin{enumerate}
    \item \textit{Transition Layer:} It maps the dimension of the output activation maps of the convolutional layers (ConvNets) to a unified dimension of spatial maps, \eg $S\times S \times D$, where $S, D$ are the dimensions of maps.
    
    \item \textit{Global Pooling Layer:} It acts globally on the whole  activation maps, \ie $S\times S \times D$, passing the important features, \ie $1\times 1 \times D$, to the prediction layer. Typical pooling operations are Average (AVG), Maximum (MAX), and Log-Sum-Exponential (LSE). 
    
    \item \textit{Prediction Layer:} It is a fully connected layer that maps the pooled features, \ie $1\times 1 \times D$  to a $C$-class one dimensional vector, \ie $1\times 1 \times C$. The weights $D\times C$ are trained using the classification loss function $\mathcal{L}_{class}$. 
\end{enumerate}   
    
Finally, the heatmaps of localized ROI $S\times S \times C$ are generated by convolving the weights of the prediction layer with the spatial maps of the transition layer. Readers are referred to~\cite{wang2017chestx} for more details. 
	% !TEX root = ../AutoClassPFF_JBHI.tex

\section{Experimental Validation}
\label{sec:Experiments}

We have designed a series of experiments to evaluate the performance of all the proposed fracture classification methods. 
First, regarding the classification itself, we distinguish three scenarios according to the number of classes considered: i) 2 classes: fracture present and normal, ii) 3 classes: fractures of type A and B and normal, and iii) 6 classes: A1, A2, A3, B1, B2, B3. We then analyze the influence of ROI localization performance on the classification accuracy. 
To this end, we evaluate the performance of supervised localization versus weakly-supervised methods. We also study different aspects of the proposed methods: first, we investigate the impact of similarity and affine transformation of our USTN model; second, we look into various weighting schemes for the auxiliary localization task in STL; finally, we evaluate three global pooling schemes. 

\textit{\textbf{Dataset Collection and Preparation.}}
Our dataset was collected by the trauma surgery department of Klinikum Rechts der Isar in Munich, Germany. It consists of 750 images from 672 patients taken with a Philips Medical Systems device with standard size of 2500x2048 px and pixel resolution varying from 0.1158 to 0.16 mm. Regarding the X-ray view, only 4\% are side-view and the remaining anterior-posterior (AP). AP images with two femurs are cropped into two, leading to one fracture and one normal case. The dataset is then composed of 1347 X-ray images each containing a single femur. 780 images contain fracture (327 from class A, and 453 from class B) while 567 are normal. Furthermore, the class distribution for 6 classes is highly imbalanced, with as little as 15 cases for class A3, and as many as 241 for class B2. We pre-process the images with histogram equalization. Clinical experts provided along with the X-ray image class annotations in the form of squared bounding boxes around the femur head.
For all our experiments, the dataset was divided into 20\% test, 10\% validation, and 70\% training data. For the two and three class scenarios, we apply data augmentation to balance the class distribution and consider $w_c=1$. For six classes, we either balance the distribution with data augmentation or use the weighted cross-entropy as loss function as defined in Eq.~\ref{eq:wce}. Data augmentation combined scaling (zoom out: 0.4-0.9, zoom in: 1.3-1.9), rotations restricted to $[-15,-5]$ and $[5,15]$ degrees and translations between the $[-1250,1250]$ pixel range in both directions ensuring the ROI does not leave the image. The number of testing images per class is 230 (115,115) for 2 classes, 175 (55,60,60) for 3 classes, and 115 (25,25,5,20,20,20) for 6 classes.

\textit{\textbf{Performance Evaluation.}}
For \textit{classification}, we report the $F_1$-score per class, and also the weighted macro average, where the weights are the support, i.e., the number of true instances for each label.
The evaluation of {\it localization} is based on the comparison of the ground truth and predicted bounding boxes. We obtain a predicted bounding box in the STN-based models by applying the transformation parameters to a canonical grid. For aff-STN, the transformed grid may not be axis-aligned, so we approximate the predicted bounding box to the corresponding square. For the other two models, STL and Global Pooling, the bounding boxes are generated from the heatmaps following the procedure in~\cite{wang2017chestx}. First, the class activation maps are resized to the input layer size with bilinear interpolation. Then, the intensities are normalized to $[0, 255]$. Finally, thresholding is applied to keep values between $\{60,180\}$. The bounding box covers the isolated regions left in the binary map. 

As localization evaluation measures, we report the \textit{mean Average Precision (mAP)} which is based on the \textit{Intersection over Union (IoU)}. 
To define the mAP, we consider several thresholds $T \in \{0.1,0.25,0.5,0.75,0.9\}$. We compare the IoU to each of the thresholds and consider the sample as TP if $ IoU > T$, otherwise is counted as FP. With this TP and FP we compute an Average Precision per threshold, we then report the \textit{mean Average Precision (mAP)}  over the thresholds.
For the anomaly detection task, we only evaluate the images from the test set that contain fractures.

\textit{\textbf{Implementation.}}
The networks were trained on a Linux-based system, with 16GB RAM, Intel(R) Xeon(R) CPU @ 3.50GHz and 64 GB GeForce GTX 1080 graphics card. All models were implemented using TensorFlow\footnote{\url{https://www.tensorflow.org/}} and initialized with pre-trained weights on ImageNet\footnote{www.image-net.org/challenges/LSVRC/}. If not otherwise stated, all the models were trained with SGD optimization algorithm with a momentum of 0.9. The batch size was set to 64 and the details of the learning rate initialization and decay are specified in Table~\ref{table:implementation}. The models were trained during 80 epochs (except Sup. Loc. for 200 epochs) and they were tested at minimum validation loss.

% Table: implementation details
\begin{table}[]
\centering
\scriptsize
\caption{Learning rate details (value and decay) of the models}. 
\vspace{-0.2cm}
\label{table:implementation}
\resizebox{0.49\textwidth}{!}{
\setlength\extrarowheight{5pt} 
\renewcommand{\arraystretch}{0.7}
\begin{tabular}{|L{2cm}|C{2cm}|C{1.45cm}|C{1.1cm}|C{2.2cm}|}
\hline
Model & Network Arch. & Learning Rate & Decay by & Decay after (epochs) \\ 
\hline
LBM, UBM, \newline USTN Classif. Net. & ResNet-50~\cite{he2016deep} & $1 \times 10^{-2}$ & 0.5 & 10 \\
\hline
Sup. Loc. & AlexNet~\cite{alexnet:nips2012} & $1 \times 10^{-8}$ & - & -  \\ 
\hline
USTN - Loc. Net. \& ST & AlexNet~\cite{alexnet:nips2012} \& STN~\cite{jaderberg2015spatial} & $1 \times 10^{-6}$ & 0.5 & 25 \\
\hline
STL & STL~\cite{hwang2016self} & $1 \times 10^{-2}$ & 0.1 & 40  \\ 
\hline
Global Pooling & ResNet-50~\cite{he2016deep} & $1 \times 10^{-3}$ & - & - \\ 
\hline
\end{tabular}
}		
\vspace{-0.2cm}	
\end{table}
\vspace{-0.5cm}	
		
	\section{Results}
	\label{sec:Results}
%%%%%%%%%%%%%%%%%%%%%%%%%%%%%%%%%%%%%%%%%%%%%%%%%%%%%%%%%%%%%%%%%%%%%%
% Table: NEW LBM, Sup.+UBM, UBM
\begin{table}[t]
\scriptsize
\centering
%\caption{Classification performance ($F_1$-score) for the 3 scenarios.}
\caption{Classification performance: $F_1$-score for the 3 scenarios. We highlight in bold the best results for each class.} %(Fig.~\ref{fig:aotree})
\vspace{-0.2cm}
\label{table:sup-loc-clas-2}
\def\colw{.2cm}
\resizebox{0.49\textwidth}{!}{  
\renewcommand{\arraystretch}{0.6}
\begin{tabular}{|L{1.4cm}|C{0.4cm}|C{\colw}C{\colw}C{\colw}C{0.3cm}|C{\colw}C{\colw}C{\colw}C{\colw}C{\colw}C{\colw}C{0.3cm}|}
\hline
& Fract. &  A\phantom{1} & B\phantom{1} & N\phantom{1} & Avg. & A1 & A2 & A3 & B1 & B2 & B3 & Avg. \\
\hline
\begin{tabular}{@{}l@{}} LBM\textsubscript{CE} \\ LBM\textsubscript{WCE} \end{tabular} &  \begin{tabular}{@{}c@{}} 0.83 \\ \phantom{0.00} \end{tabular} & \begin{tabular}{@{}c@{}} 0.76 \\ \phantom{0.00} \end{tabular} & \begin{tabular}{@{}c@{}} 0.81 \\ \phantom{0.00} \end{tabular} & \begin{tabular}{@{}c@{}} 0.89 \\ \phantom{0.00} \end{tabular} & \begin{tabular}{@{}c@{}} 0.82 \\ \phantom{0.00} \end{tabular} & \begin{tabular}{@{}c@{}} 0.26 \\ 0.22  \end{tabular} &
\begin{tabular}{@{}c@{}} 0.73 \\ {\bf 0.76} \end{tabular} &
\begin{tabular}{@{}c@{}} 0.00 \\ 0.00  \end{tabular}  &
\begin{tabular}{@{}c@{}} 0.40 \\ 0.56 \end{tabular}  & 
\begin{tabular}{@{}c@{}} 0.35 \\ 0.54 \end{tabular}  & 
\begin{tabular}{@{}c@{}} 0.34 \\ 0.53 \end{tabular}   &
\begin{tabular}{@{}c@{}} 0.41 \\ 0.49 \end{tabular} \\
\hline
a-STN &  0.92 &  0.74 & 0.82 & 0.88 & 0.82 &  0.07 & 0.71 & 0.00 & 0.32 & 0.36 & 0.16 & 0.32 \\
\hline
aff-STN &  0.90 &  0.80 & 0.87 & 0.86 & 0.84 & 0.33 & 0.75 & 0.00 & 0.50 & 0.47 & 0.43 & 0.48 \\
\hline
\begin{tabular}{@{}l@{}} Sup.+UBM\textsubscript{CE} \\ Sup.+UBM\textsubscript{WCE} \end{tabular} &
			\begin{tabular}{@{}c@{}} 0.93 \\ \phantom{0.00} \end{tabular} & 
			\begin{tabular}{@{}c@{}} 0.81 \\ \phantom{0.00} \end{tabular} & 
			\begin{tabular}{@{}c@{}} 0.83 \\ \phantom{0.00} \end{tabular} & 
			\begin{tabular}{@{}c@{}} {\bf 0.91} \\ \phantom{0.00} \end{tabular} & 
			\begin{tabular}{@{}c@{}} 0.85 \\ \phantom{0.00} \end{tabular} &
			\begin{tabular}{@{}c@{}} 0.50 \\ 0.47 \end{tabular} &
			\begin{tabular}{@{}c@{}} 0.71 \\ {\bf 0.76} \end{tabular} &
			\begin{tabular}{@{}c@{}} 0.00 \\ 0.00 \end{tabular} &
			\begin{tabular}{@{}c@{}} {\bf 0.63} \\ 0.55 \end{tabular} &
			\begin{tabular}{@{}c@{}} 0.40 \\ {\bf 0.49} \end{tabular} &
			\begin{tabular}{@{}c@{}} {\bf 0.63} \\ 0.38 \end{tabular} &
			\begin{tabular}{@{}c@{}} \textbf{0.55} \\ 0.51 \end{tabular} \\
			% &  &  &  &  &  &  &  &  &  &  &  & \\
			\hline
\begin{tabular}{@{}l@{}} UBM\textsubscript{CE} \\ UBM\textsubscript{WCE} \end{tabular} &
			\begin{tabular}{@{}c@{}} {\bf 0.94} \\ \phantom{0.00} \end{tabular} & 
			\begin{tabular}{@{}c@{}} {\bf 0.85} \\ \phantom{0.00} \end{tabular} & 
			\begin{tabular}{@{}c@{}} {\bf 0.88} \\ \phantom{0.00} \end{tabular} & 
			\begin{tabular}{@{}c@{}} 0.88 \\ \phantom{0.00} \end{tabular} & 
			\begin{tabular}{@{}c@{}} \textbf{0.87} \\ \phantom{0.00} \end{tabular} &
			\begin{tabular}{@{}c@{}} 0.19 \\ 0.18 \end{tabular} &
			\begin{tabular}{@{}c@{}} {\bf 0.76} \\ 0.71 \end{tabular} &
			\begin{tabular}{@{}c@{}} {\bf 0.33} \\ 0.00 \end{tabular} &
			\begin{tabular}{@{}c@{}} 0.55 \\ 0.52 \end{tabular} &
			\begin{tabular}{@{}c@{}} 0.47 \\ 0.45 \end{tabular} &
			\begin{tabular}{@{}c@{}} 0.47 \\ 0.24 \end{tabular} &
			\begin{tabular}{@{}c@{}} 0.48 \\ 0.40 \end{tabular} \\
			\hline
		\end{tabular}
	}
	\vspace{-0.55cm}
	\end{table}
	%%%%%%%%%%%%%%%%%%%%%%%%%%%%%%%%%%%%%%%%%%%%%%%%%%%%%%%%%%%%%%%%%%%%%%
	
	\subsubsection{Automatic Classification}  
We present here our baseline classification results for the 2, 3 and 6 class scenarios. We report the classification metrics for LBM, UBM as well as for the predicted ROI produced by the supervised localization network. From the classification results in Table \ref{table:sup-loc-clas-2}, we confirm that the supervised localization network is able to predict reasonable ROIs in all three scenarios. For instance, for 2 classes, $F_1$-score is improved by 10\% \wrt the LBM and reaches a performance comparable to the UBM. Similarly, for 3 classes, the improvement is of 3\%. For 6 classes, the combination of supervised localization and UBM improves the $F_1$-score result by 14\%. We also show that a ConvNet architecture such as AlexNet \cite{alexnet:nips2012} is able to achieve acceptable results when regressing the bounding box for the femur head with a mAP of 0.71. 
		
\subsubsection{Weakly-Supervised Models} \hfill\\
\indent \textit{\textbf{Attention Model.}}	
First, we compare the performance of the USTN model  with our baselines and analyze the influence of parameterizing USTN's warping transformation with 3 or 6 parameters. Table~\ref{table:sup-loc-clas-2} shows the performance of both a-STN, and aff-STN. The results of a-STN are on par with those in our previous work~\cite{kazi2017automatic}. It should be noted that both the localization and classification networks chosen in~\cite{kazi2017automatic} were GoogleNet (inception v3) \cite{inception:cvpr2015} and the STN was modified by including a differentiable sigmoid layer. Here, we use instead a ResNet and no modification.

	We observe that for binary classification a-STN improves the $F_1$-score by 12\% (2.5\% in \cite{kazi2017automatic}), while its performance is similar for 3 classes and worse for 6 classes (decreased by 22\% here, 20\% in \cite{kazi2017automatic}). aff-STN has a  similar behavior for 2 and 3 classes, while for the 6 classes scenario, the $F_1$-score surprisingly improves by 17\%, reaching the performance of the UBM based on supervised localization. 	

\begin{figure}
	\vspace{-0.6cm}
		\centering 
		\includegraphics[width=0.8\linewidth]{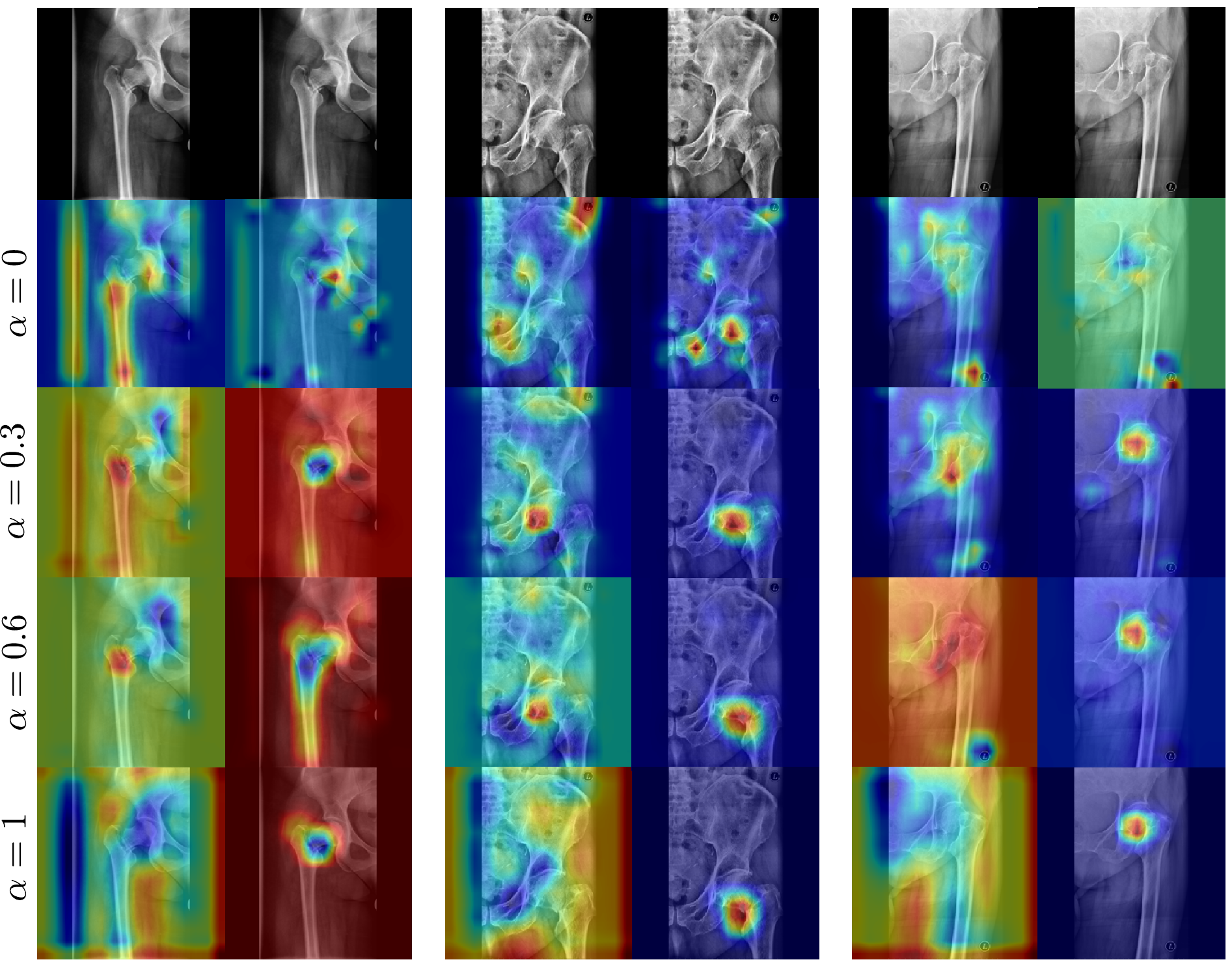}
		\caption{Visualization of the activations for 6 classes of STL. In each of the 3 columns we show an example of classes A1, B2 and B3, respectively. Within each column, left: normal cross entropy, right: weighted cross entropy.}{}
		\label{fig:stl-vis6w}
			\vspace{-0.5cm}
	\end{figure}

\textit{\textbf{Self-Transfer Learning.}}	
For the STL method, we evaluate different values of $\alpha$ including the extreme cases $\alpha=\{0,1\}$ where either the localization or classification branch are not considered during training. The results reported in Table~\ref{table:results-classif-stl-full} show an improvement of $F_1$-score \wrt the baseline ($\alpha=0$) by 6\%, 4\% and 11\% for 2, 3 and 6 classes, respectively. For localization, mAP is almost similar to the baseline for 2 classes, while it shows better performance by 6\% and 12\% for 3 and 6 classes, respectively. Our results prove that localization task ($\alpha>0$) positively influences the classification accuracy and  leads to meaningful and explainable heatmaps (\cf Fig.~\ref{fig:stl-vis6w}).
	The best performance on both classification and localization metrics is obtained at $\alpha=0.6$ and with weighted cross entropy loss (\cf Fig.~\ref{fig:stl-vis6w}).
	
	%%%%%%%%%%%%%%%%%%%%%%%%%%%%%%%%%%%%%%%%%%%%%%%%%%%%%%%%%%%%%%%%%%%%%%
	% Table: new Full Results STL
	% Classification performance
	\begin{table}[t]
		\scriptsize
		\centering
%	    \subfloat[][AAAAAAAAAAA.]
	%\caption{$F_1$-score for different $\alpha$ schemes and for the 3 scenarios.} 
		\caption{Comparison of different $\alpha$ schemes for the 3 scenarios, we report $F_1$-score for classification performance.}
		\vspace{-0.3cm}
		\def\colw{.2cm}
		\resizebox{0.49\textwidth}{!}{ 
	    \renewcommand{\arraystretch}{0.8} 
		\begin{tabular}{|L{1.3cm}|C{0.3cm}|C{\colw}C{\colw}C{\colw}C{0.3cm}|C{\colw}C{\colw}C{\colw}C{\colw}C{\colw}C{\colw}C{0.3cm}|}
			% \hline
			%Classes & 2  &  \multicolumn{4}{c|}{3} & \multicolumn{7}{c|}{6}\\
			\hline
			STL\phantom{123456789}  & Frac. &  A\phantom{1} & B\phantom{1} & N\phantom{1} & Avg. & A1 & A2 & A3 & B1 & B2 & B3 & Avg. \\
			\hline
			\begin{tabular}{@{}l@{}} $\alpha=0.0$, \textsubscript{CE} \\ $\alpha=0.0$, \textsubscript{WCE} \end{tabular} &  \begin{tabular}{@{}c@{}} 0.84 \\ \phantom{0.00} \end{tabular} & \begin{tabular}{@{}c@{}} 0.74 \\ \phantom{0.00} \end{tabular} & \begin{tabular}{@{}c@{}} {\bf 0.87} \\ \phantom{0.00} \end{tabular} & \begin{tabular}{@{}c@{}} 0.80 \\ \phantom{0.00} \end{tabular} & \begin{tabular}{@{}c@{}} 0.77 \\ \phantom{0.00} \end{tabular} & \begin{tabular}{@{}c@{}} 0.06 \\ 0.20  \end{tabular} &
			\begin{tabular}{@{}c@{}} 0.56 \\ 0.67 \end{tabular} &
			\begin{tabular}{@{}c@{}} 0.00 \\ 0.00  \end{tabular}  &
			\begin{tabular}{@{}c@{}} 0.16 \\ 0.37 \end{tabular}  & 
			\begin{tabular}{@{}c@{}} 0.26 \\ {\bf 0.49} \end{tabular}  & 
			\begin{tabular}{@{}c@{}} 0.20 \\ 0.08 \end{tabular}   &
			\begin{tabular}{@{}c@{}} 0.24 \\ 0.35 \end{tabular} \\
			\hline
			\begin{tabular}{@{}l@{}} $\alpha=0.3$, \textsubscript{CE} \\ $\alpha=0.3$, \textsubscript{WCE} \end{tabular} &  \begin{tabular}{@{}c@{}} 0.89 \\ \phantom{0.00} \end{tabular} & \begin{tabular}{@{}c@{}} {\bf 0.78} \\ \phantom{0.00} \end{tabular} & \begin{tabular}{@{}c@{}} 0.82 \\ \phantom{0.00} \end{tabular} & \begin{tabular}{@{}c@{}} 0.85 \\ \phantom{0.00} \end{tabular} & \begin{tabular}{@{}c@{}} 0.71 \\ \phantom{0.00} \end{tabular} & \begin{tabular}{@{}c@{}} 0.22 \\ 0.26  \end{tabular} &
			\begin{tabular}{@{}c@{}} 0.70 \\ 0.68 \end{tabular} &
			\begin{tabular}{@{}c@{}} 0.00 \\ 0.00  \end{tabular}  &
			\begin{tabular}{@{}c@{}} 0.37 \\ {\bf 0.44} \end{tabular}  & 
			\begin{tabular}{@{}c@{}} 0.35 \\ 0.42 \end{tabular}  & 
			\begin{tabular}{@{}c@{}} 0.21 \\ 0.42 \end{tabular}   &
			\begin{tabular}{@{}c@{}} 0.36 \\ 0.43 \end{tabular} \\
			\hline   
			\begin{tabular}{@{}l@{}} $\alpha=0.6$, \textsubscript{CE} \\ $\alpha=0.6$, \textsubscript{WCE} \end{tabular} &  \begin{tabular}{@{}c@{}} \textbf{0.90} \\ \phantom{0.00} \end{tabular} & \begin{tabular}{@{}c@{}} 0.74 \\ \phantom{0.00} \end{tabular} & \begin{tabular}{@{}c@{}} 0.82 \\ \phantom{0.00} \end{tabular} & \begin{tabular}{@{}c@{}} {\bf 0.88} \\ \phantom{0.00} \end{tabular} & \begin{tabular}{@{}c@{}} \textbf{0.81} \\ \phantom{0.00} \end{tabular} & \begin{tabular}{@{}c@{}} 0.07 \\ {\bf 0.37}  \end{tabular} &
			\begin{tabular}{@{}c@{}} 0.71 \\ {\bf 0.73} \end{tabular} &
			\begin{tabular}{@{}c@{}} 0.00 \\ 0.00  \end{tabular}  &
			\begin{tabular}{@{}c@{}} 0.32 \\ 0.40 \end{tabular}  & 
			\begin{tabular}{@{}c@{}} 0.36 \\ 0.43 \end{tabular}  & 
			\begin{tabular}{@{}c@{}} 0.16 \\ {\bf 0.52} \end{tabular}   &
			\begin{tabular}{@{}c@{}} 0.32 \\ \textbf{0.47} \end{tabular} \\
			\hline
			\begin{tabular}{@{}l@{}} $\alpha=1.0$, \textsubscript{CE}  \\ $\alpha=1.0$, \textsubscript{WCE} \end{tabular} &  \begin{tabular}{@{}c@{}} 0.44 \\ \phantom{0.00} \end{tabular} & \begin{tabular}{@{}c@{}} 0.48 \\ \phantom{0.00} \end{tabular} & \begin{tabular}{@{}c@{}} 0.03 \\ \phantom{0.00} \end{tabular} & \begin{tabular}{@{}c@{}} 0.00 \\ \phantom{0.00} \end{tabular} & \begin{tabular}{@{}c@{}} 0.17 \\ \phantom{0.00} \end{tabular} & \begin{tabular}{@{}c@{}} 0.00 \\ 0.07  \end{tabular} &
			\begin{tabular}{@{}c@{}} 0.00 \\ 0.00 \end{tabular} &
			\begin{tabular}{@{}c@{}} 0.09 \\ {\bf 0.10}  \end{tabular}  &
			\begin{tabular}{@{}c@{}} 0.29 \\ 0.00 \end{tabular}  & 
			\begin{tabular}{@{}c@{}} 0.00 \\ 0.27 \end{tabular}  & 
			\begin{tabular}{@{}c@{}} 0.00 \\ 0.11 \end{tabular}   &
			\begin{tabular}{@{}c@{}} 0.05 \\ 0.09 \end{tabular} \\
			\hline
		\end{tabular}
	}
	\vspace{-0.4cm}
		\label{table:results-classif-stl-full}
	\end{table}
	%%%%%%%%%%%%%%%%%%%%%%%%%%%%%%%%%%%%%%%%%%%%%%%%%%%%%%%%%%%%%%%%%%%%%%
	%%w
	
%%%%%%%%%%%%%%%%%%%%%%%%%%%%%%%%%%%%%%%%%%%%%%%%%%%%%%%%%%%%%%%%%%%%%%
% Table Summary Results STL
\begin{table}[]
\centering
\scriptsize
\caption{Comparison of different $\alpha$ schemes for the 3 scenarios, we report mAP for localization performance.}
\vspace{-0.3cm}
\label{table:stl}
\def\colw{1.5cm}
\resizebox{0.49\textwidth}{!}{ 
\begin{tabular}{|L{\colw}|C{\colw}C{\colw}C{2.4cm}|}
\hline
STL & 2 Classes & 3 Classes & 6 Classes (CE / WCE) \\
\hline
$\alpha = 0$ & \textbf{0.28} & 0.27 & 0.26 / 0.23 \\
$\alpha = 0.3$ & 0.26 & 0.30 & 0.32 / 0.32 \\
$\alpha = 0.6$ & 0.27 & \textbf{0.33} & \textbf{0.35} / \textbf{0.35} \\
$\alpha = 1$ & 0.19 & 0.24 & 0.23 / 0.28 \\
\hline
\end{tabular}
}	
\vspace{-0.5cm}		
\end{table}
	
	%%%%%%%%%%%%%%%%%%%%%%%%%%%%%%%%%%%%%%%%%%%%%%%%%%%%%%%%%%%%%%%%%%%%%%
	% Figure: Visualize activations STL - 6 classes COMBINED
	%%%%%%%%%%%%%%%%%%%%%%%%%%%%%%%%%%%%%%%%%%%%%%%%%%%%%%%%%%%%%%%%%%%%%%
	
\begin{figure}
	\centering 
	\includegraphics[width=0.8\linewidth]{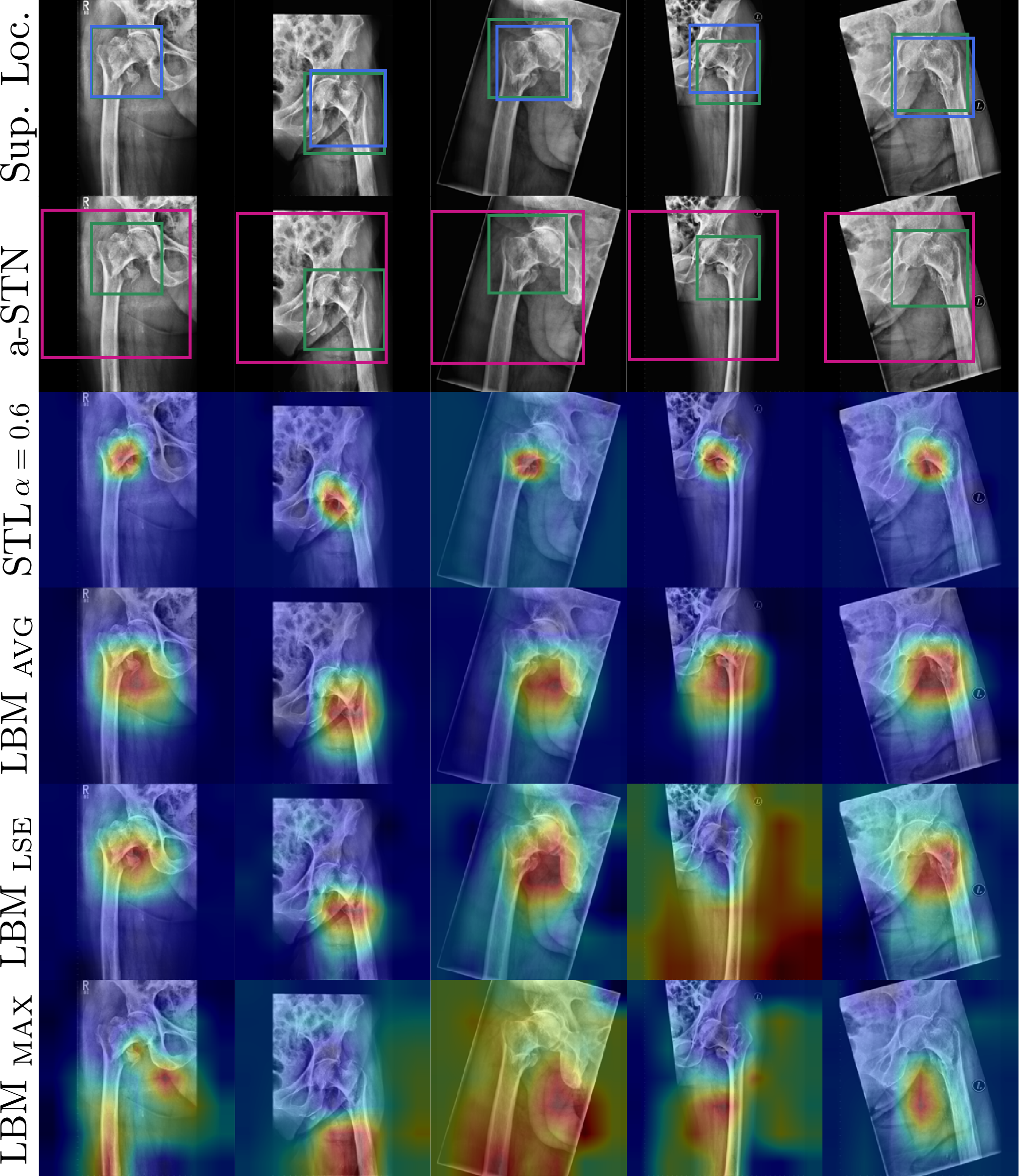}
	\caption{
Bounding box and heatmap predictions by the different methods. The green bounding boxes correspond to ground truth localization.}{}
	~\label{fig:activationmaps}
	\vspace{-0.7cm}
\end{figure}

\textit{\textbf{Global Pooling:}}	
The activations based on guided localization as shown in Fig.~\ref{fig:activationmaps} are given as input to the pooling layer. Better localization produce high activations at the ROI, hence AVG pooling catches this information better than MAX pooling which stays sensitive to arbitrary activations due to artifacts in the image. LSE is mainly designed for  multiple anomalies co-exiting in same image, hence is not powerful with dataset like ours with one anomaly per image~\cite{wang2017chestx}.
These results are consistent with the qualitative localization results depicted in Fig.~\ref{fig:activationmaps}.
    
We evaluate the robustness of different pooling schemes on our baseline models LBM and UBM, as well as a-STN, aff-STN, in order to analyze the activations through the network. These models were run using Adagrad optimizer and an initial learning rate of $1 \times 10^{-3}$. Our result from Table~\ref{table:classif-pools} confirm the best performance of AVG pooling. 
However LBM and UBM show reduced performance with AVG pooling as no localization network is used in both the cases to provide activations at ROI.
	
	%%%%%%%%%%%%%%%%%%%%%%%%%%%%%%%%%%%%%%%%%%%%%%%%%%%%%%%%%%%%%%%%%%%%%%
	% Table: Pooling Schemes
	% Classification & Localization
	\begin{table}[t]
		\scriptsize
		\centering
		%\caption{$F_1$-score (Classif.) and mAP (Localiz.) under various poolings.}
		\caption{Classification: $F_1$-score (top) and localization: mAP (bottom) performance under various global poolings.}
		\vspace{-0.2cm}
		\def\colw{.28cm} 
		\resizebox{0.49\textwidth}{!}{
	    \renewcommand{\arraystretch}{0.25}
		\begin{tabular}{|L{0.815cm}|C{\colw}C{\colw}C{\colw}|C{\colw}C{\colw}C{\colw}|C{\colw}C{\colw}C{\colw}|}
			\hline
			& \multicolumn{3}{c|}{2 classes} & \multicolumn{3}{c|}{3 classes} & \multicolumn{3}{c|}{6 classes}  \\
			\hline
			Classif.\phantom{12} & MAX & LSE & AVG & Max & LSE & AVG & MAX & LSE & AVG\\
			\hline
			LBM & 0.68 & 0.87 & \textbf{0.89} & 0.63 & 0.63 & \textbf{0.78} & 0.15 & 0.31 & \textbf{0.33} \\ 
			a-STN  & 0.84 & 0.84 & \textbf{0.85} & \textbf{0.78} & 0.76 & 0.73  & 0.26 & \textbf{0.37} & \textbf{0.37} \\ 
			aff-STN  & 0.84 & \textbf{0.85} & 0.84          & 0.74          & 0.76 & \textbf{0.79} & 0.27 & \textbf{0.36} & \textbf{0.36} \\
			UBM  & 0.39 & 0.88          & \textbf{0.93} & 0.36          & 0.81 & \textbf{0.84} & 0.18 & 0.36          & \textbf{0.46}          \\ \hline
			Localiz.\phantom{12} & MAX & LSE & AVG & MAX & LSE & AVG & MAX & LSE & AVG\\
			\hline
			LBM & 0.23 & 0.23 & \textbf{0.40} & 0.31 & 0.26 & \textbf{0.47} & 0.23 & 0.23 & \textbf{0.37} \\
			a-STN, aff-STN & 0.32 & 0.32 & 0.32 & 0.32 & 0.32  & 0.32 & 0.32  & 0.32 & 0.32\\   
			\hline
		\end{tabular}}
		\vspace{-0.2cm}
		\label{table:classif-pools}
	\end{table}
	%%%%%%%%%%%%%%%%%%%%%%%%%%%%%%%%%%%%%%%%%%%%%%%%%%%%%%%%%%%%%%%%%%%%%%
	
%% Compiled figure	
\vspace{-0.3cm}

	\section{Discussion and Conclusion}
\label{sec:Discussion}
In this work, we have presented an analysis of suitable techniques for proximal femur fracture classification. We put forward various models to perform the task, considering two different approaches. 

First, we have used the spatial annotations provided by the experts to improve our baseline performance and build a bounding box predictor. We have verified the importance of localization in this task, by comparing against the lower- and upper-bound models. As hypothesized, a closer look onto the fractured bone does improve the performance of the models in every scenario. All our models give good results for 2 and 3 classes.
However, the 6 class scenario is really challenging and our results are slightly below the inter-observer variability of 66-71\%.
We believe that increasing the size of the dataset is necessary, as for some classes it reaches a critical number (\eg 15 for class A3), which is far from representative of the true intra-class variation.  Other possibilities to deal with the 6 classes are the use of more complex architectures like a ConvNet cascade to fetch deeper features, or different learning strategies such as curriculum learning~\cite{bengio2009curriculum} that gradually makes the task harder.

In a second series of approaches, we have explored supervised and weakly-supervised methods to the localization task, with the motivations of reducing the  annotation efforts and  improving the LBM results. 
First, we confirm that computed bounding boxes with our Sup. Loc. model are beneficial for the performance of the 6-class scenario.
Regarding weak-supervision, we conclude from Tab.~\ref{table:sup-loc-clas-2}, that weakly-supervised methods are useful, in particular, for applications with noisy annotations such as ours.  Moreover, we get this localization information for free.
In any case, the prediction of bounding boxes has a clear added value as the decisions can be better explained. For example, the visualization of the score maps in Fig.~\ref{fig:stl-vis6w} helps us to better understand  the behavior of the networks and analyze the features that are given more importance.  This interpretability is useful despite the fact that the best reached mAP for localization was rather low 0.47. We believe that the used localization metric, based on the IoU, is  too strongly affected by errors in scale of the bounding box, while scale is not that relevant to the classification.

The weakly-supervised models implicitly learn the localization information through back propagation of the classification loss either in a sequential (USTN) or parallel (STL) manner. In case of USTN, the initial steps of the optimization are very critical. We initialize the transformation to take a crop from the center of the image  with scale equal to 1. This crops the whole image as a ROI, such that  the classification network is equivalent to the LBM. In our previous work~\cite{kazi2017automatic}, we considered the use of a differentiable sigmoid layer on top of ST module. The sigmoid restricted the space of the transformation parameters. Moreover, the gradients were clipped and set to zero whenever $|t_r|,|t_c|\, \text{or}\, |s| > 1$. In this paper, we have found that fine-tuning the learning rate in a layer-wise fashion performs similarly without restricting the values of the parameters. 

Regarding the complexity of the models, we note that retrieving the heatmaps from the LBM, does not add any extra complexity. Conversely,  STL contains an extra localization branch, and for STN, an additional localization network. Regarding convergence, UBM and LBM converge faster than the compared methods, as composed of a single network. The model that takes longer to converge is aff-STN. We believe that learning filters in a sequential manner, instead of parallel (like STL), affects adversely the speed of convergence. Overall, all the models except Sup.~+~UBM, take on average 25 minutes to train.

To sum up, the localization task is important for the accurate fine-grained classification of fracture types according to the AO standard, as shown by the 9\% of average improvement of our UBM w.r.t. the LBM. In absence of bounding box annotations, weakly-supervised methods provide free localization information, while boosting the classification performance, for instance by 7\% for STL w.r.t. LBM. Moreover, weakly-supervised localization improves the interpretability of the decisions. We hope that our analysis and results impact other hierarchical classification tasks coming from different medical image applications.
\vspace{-0.4cm}	
	
	% or
	
	%\appendices
	%\section{}
	\section*{Acknowledgment}
	The authors would like to thank our clinical partners, in particular, Ali Deeb, PD,  Dr. med. Marc Beirer, and Fritz Seidl, MA, MBA. for their support during the work.  %This work was supported -- in part -- by the European Union’s Horizon 2020 research and innovation programme under the Marie Sklowdowska-Curie grant agreement No. 713673. Amelia Jim\'enez-S\'anchez has received financial support through the “la Caixa” INPhINIT Fellowship Grant for Doctoral studies at Spanish Research Centres of Excellence, “la Caixa” Banking Foundation, Barcelona, Spain. The authors thank Nvidia for the GeForce GTX 1080 GPU donation.
	\vspace{-0.2cm} 
	\ifCLASSOPTIONcaptionsoff
	\newpage
	\fi
	
	%\bibliographystyle{IEEEtran}
	%\bibliographystyle{splncs03}
	
	\bibliographystyle{IEEEtran}
	\bibliography{biblio}	
	
\end{document}